\documentclass[]{ceurart}

\usepackage{listings}
\usepackage{xcolor}
\usepackage{array}
\usepackage{float}
\usepackage{longtable}
\usepackage{caption}
\begin{document}

\copyrightyear{2025}
\copyrightclause{Copyright for this paper by its authors. Use permitted under Creative Commons License Attribution 4.0 International (CC BY 4.0).}

\conference{IberLEF 2025, September 2025, Zaragoza, Spain}

\title{HULAT-UC3M @ CLEARS2025 Subtask~1: Prompt-Based Simplification for Plain Language using Spanish Language Models}

\address[1]{Universidad Carlos III de Madrid, Av. Universidad, 30, Legan{\'e}s, 28911, Spain}

\author[1]{Lourdes Moreno}[%
orcid=0000-0002-9021-2546,
email=lmoreno@inf.uc3m.es,
url=https://hulat.inf.uc3m.es/,
]

\author[1]{Jesus M. Sanchez-Gomez}[%
orcid=0000-0002-6415-7467,
email=jesusmsa@inf.uc3m.es,
url=https://hulat.inf.uc3m.es/,
]

\author[1]{Marco Antonio Sanchez-Escudero}[%
orcid=0009-0001-8163-5440,
email=marcoasa@inf.uc3m.es,
url=https://hulat.inf.uc3m.es/,
]

\author[1]{Paloma Mart{\'i}nez}[%
orcid=0000-0003-3013-3771,
email=paloma.martinez@uc3m.es,
url=https://hulat.inf.uc3m.es/,
]

\cortext[1]{Corresponding author.}
\fntext[1]{These authors contributed equally.}

\begin{abstract}
This paper describes the participation of HULAT-UC3M in CLEARS 2025 Subtask~1: Adaptation of Text to Plain Language (PL) in Spanish. We explored strategies based on models trained on Spanish texts, including a zero-shot configuration using prompt engineering and a fine-tuned version with Low-Rank Adaptation (LoRA). Different strategies were evaluated on representative internal subsets of the training data, using the official task metrics, cosine similarity (SIM) and the Fern{\'a}ndez-Huerta readability index (FH) to guide the selection of the optimal model and prompt combination. The final system was selected for its balanced and consistent performance, combining normalization steps, the RigoChat-7B-v2 model, and a dedicated PL-oriented prompt. It ranked first in semantic similarity (SIM = 0.75), however, fourth in readability (FH = 69.72). We also discuss key challenges related to training data heterogeneity and the limitations of current evaluation metrics in capturing both linguistic clarity and content preservation.
\end{abstract}


\maketitle

\section{Introduction}\label{sec:introduction}

Automatic text simplification using Natural Language Processing (NLP) methods is gaining increasing relevance in domains such as healthcare, public services, and finance. These tools aim to improve readability and comprehension, particularly for people with cognitive impairments or reading difficulties. However, to ensure real-world and social impact, simplification systems must not only align with international standards that guarantee the right to accessible information, but also involve end users in the NLP process itself. Within this framework, there are the initiatives of Plain Language (PL) and Easy Reading (ER), which promote the adaptation of texts to user needs and capabilities, especially people with intellectual disabilities or difficulties in reading comprehension, particularly in essential domains such as public administration, healthcare, education, and other key areas. The main international reference is ISO~24495-1:2023, which defines general principles applicable to any language and domain, stating that texts should be easy to find, understand, use, and evaluate by their target audience \cite{ISO2449512023}.

PL and ER initiatives constitute the methodological basis for textual simplification, as both aim to optimize vocabulary, sentence structure, and visual design to enhance message comprehension. UNE~153101:2018~EX in Spain, for example, provides guidelines for producing and validating documents in ER that explicitly require validation by people with intellectual disabilities \cite{UNE1531012018}. Although PL and ER share the general objective of making the text understandable, there are important differences. PL is aimed at a broad audience, seeking to eliminate technicalities and complex expressions. Its principles are based on short sentences, the use of inclusive and everyday language, and a structure that allows the key information to be located. It usually employs direct style, active voice, and segmentation into short paragraphs for easy reading. However, ER is specifically designed for people with intellectual disabilities and people with reading comprehension difficulties, and therefore requires even stricter guidelines for simplification. Additionally, some content presentation guidelines include bulleted lists, high-frequency words, abundant use of visual examples, and single-idea sentences, making it easier to read for people with cognitive barriers. 

The participation of HULAT-UC3M in CLEARS Subtask~1 \cite{clears_at_iberlef2025}, within the IberLEF 2025 framework \cite{iberlef2025overview}, focused on the use of Spanish generative Large Language Models (LLMs) through prompt engineering. Our goal was to explore different prompt strategies to generate high-quality PL adaptations, evaluating their effectiveness in balancing semantic fidelity and readability. 

\section{Dataset and Evaluation Metrics}\label{sec:dataset}

This section describes the training and evaluation datasets, along with the evaluation metrics provided by CLEARS Subtask~1 to guide system design.

\subsection{CLEARS Subtask~1 Dataset}\label{subsec:dataset_description}

As a first step, we explored the training dataset \cite{espinosa-zaragoza-etal-2023-automatic,botella2024clearsim} provided by the organizers of the Subtask~1 at CLEARS 2025 task \cite{clears_at_iberlef2025}. It is composed of 2,400 texts extracted from official communications published between 2022 and 2023 by various Spanish municipalities, including Madrid, Alicante, Benidorm, Orihuela, Torrevieja, and Elda. These texts include topics of public interest and administrative relevance, and constitute the base material for evaluating textual adaptation to PL (see Table~\ref{tab:dataset}).

\begin{table*}
  \caption{Statistics of provided CLEARS Subtask~1 dataset}
  \label{tab:dataset}
  \begin{tabular}{l r r r}
    \toprule
    \textbf{Description} & \multicolumn{2}{c}{\textbf{Train dataset}} & \multicolumn{1}{c}{\textbf{Test dataset}} \\
    & Original Texts &  Adaptations &  Adaptations\\
    \midrule
    No. of samples & 2,400 & 2,400 & 607  \\
    Avg. no. of words & 431.48 & 189.31 & 326.43 \\
    Avg. no. of lines & 6.63 & 5.30 & 10.77 \\
    Avg. word length (characters) & 5.11 & 4.88 & 5.19 \\
  \bottomrule
\end{tabular}
\end{table*}

The adaptation methodology or guidelines followed by the experts who created the PL versions in the training dataset were not provided. Therefore, our aim was to analyze how the adaptations have been carried out. Although our research methodology follows ISO~24495-1:2023, we did not find clear evidence of its application in the training dataset. As a result, we chose not to apply our trained models and prompt strategies and instead to adapt our approach to reflect the style of the observed adaptations. 

We found various inconsistencies in the training dataset that posed challenges for developing and evaluating our systems. For example, some adapted texts included information not present in the original (e.g., entry ID=2346), while others omitted relevant content due to excessive simplification (e.g., ID=1075). 

In the test dataset, several source entries lacked a coherent semantic unit suitable for PL adaptation, such as:
\begin{itemize}
  \item ``Las actividades veraniegas en Parques contin{\'u}an esta semana con:'' \textit{(``Summer activities in parks continue this week with:'')} (ID=542)
  \item “M{\'a}s actividades en www.agendacultural.org'' \textit{(``More activities at www.agendacultural.org'')} (ID=538)
\end{itemize}

These issues introduced ambiguity and hindered the consistent application of adaptation strategies (see Table~\ref{tab:dataset-examples} in the Appendix for representative examples).

\subsection{Evaluation Metrics}\label{subsec:metrics}

Subtask~1 at CLEARS 2025 defined two evaluation metrics to assess the quality of the PL-adapted outputs: Semantic Similarity (SIM) to measure content preservation, and the Fern{\'a}ndez-Huerta Readability Index (FH) to assess linguistic clarity in Spanish. These metrics were applied to all generated outputs in both development and official test phases.

\begin{itemize}
  \item \textbf{Semantic Similarity.} Semantic alignment between each system output and its PL reference was evaluated using cosine similarity under two complementary approaches: Bag-of-Words (BoW) and sentence embeddings. For the BoW method, we used \texttt{CountVectorizer} to obtain frequency vectors. For the embedding-based method, we used the multilingual model \texttt{paraphrase-multilingual-mpnet-base-v2} from \texttt{SentenceTransformers}. The final SIM score was computed as the average of both methods. The cosine similarity ranges from $-1$ to $1$, with higher values indicating stronger semantic overlap.
  \item \textbf{Fern{\'a}ndez-Huerta Readability Index.} This index is a standard formula for Spanish readability, based on the number of syllables per word and the number of sentences per 100 words. We implemented it in Python using the original mathematical formulation, combining the \texttt{pyphen} library for syllable segmentation and regular expressions for sentence boundary detection. The index yields a score from 0 (very difficult) to 100 (very easy), providing a quantifiable estimate of textual clarity.
\end{itemize}

\section{Proposed Architecture}\label{sec:architecture}

The architecture proposed by HULAT-UC3M was based on the application of prompt engineering to pre-trained transformer models with Spanish texts, in order to analyze the quality of texts adapted to PL. The objective was to maximize the metrics used in the shared task: Cosine similarity measure~\cite{salton1988term} and Fern{\'a}ndez-Huerta readability index~\cite{fernandezhuerta1959}.

Concerning the generative model to be integrated, there is a wide variety of generative models, both encoder-decoder and decoder types, such as Mistral-7B-Instruct~\cite{mistralai_2023}, Gemma~\cite{google_2024}, and StableLM~\cite{stabilityai_2024}. However, they did not provide the level of linguistic quality or instructional control required for adapting texts to PL in Spanish. Processing was performed on infrastructure equipped with 23~GB of GPU memory and sufficient disk storage.

After this review and prioritizing models trained with Spanish texts, Salamandra~\cite{gonzalezagirre2025salamandratechnicalreport} and RigoChat~\cite{gomez2025rigochat2adaptedlanguage,instituto_de_ingenieria_del_conocimiento_2025} were selected as the most promising for text adaptation to PL. The Salamandra family includes Salamandra-7B-Instruct, which has been fine-tuned through instruction tuning, enabling it to follow instructions in generative tasks such as simplification, summarization, or rewriting~\cite{gonzalezagirre2025salamandratechnicalreport,bsc_2024}. RigoChat, developed by the Instituto de Ingenier{\'i}a del Conocimiento (IIC), is a generative LLM specialized in Spanish, trained with an instruction-tuning strategy focused on useful tasks for Spanish-speaking users. One of its key features is that it was trained under limited computational conditions, which demonstrates its applicability in resource-constrained environments. Moreover, RigoChat has been fine-tuned to maintain a balance between accuracy, clarity, and instructional control, which are fundamental aspects for the task of text simplification and adaptation to PL ~\cite{iic_rigochat_libera_2024,iic_rigochat_practico_2024}. Additionally, we explored fine-tuning strategies like Low-Rank Adaptation (LoRA) in Spanish generative models. These strategies, together with the preprocessing steps and the system variants described below, are detailed in the next sections.

\subsection{Text Normalization}\label{subsec:normalization}

To adapt the output of the model to the type of adaptations observed in the training data, a set of filters and rules (implemented in \textit{Python} using custom \textit{regex}) were applied as pre and postprocessing. These processes ensured that the output of the model was consistent with the analysis of the training data. Examples of representative transformations include the following.
\begin{itemize}
  \item Dates and times: expressions such as ``20:00 hours'' are normalized to more natural forms such as ``at 8 in the evening''.
  \item Conversion of monetary amounts: numeric expressions like ``13.50 EUR'' are transformed into simpler forms, for example, ``13 with 50 euros''.
  \item Protection and normalization: special numeric elements such as percentages, years, date ranges, and large numbers are processed to prevent misinterpretations with the model.
\end{itemize}

\subsection{Prompt Design}\label{subsec:prompts}

To identify the most effective configuration for Subtask~1, we evaluated both Salamandra and RigoChat models using various prompt strategies. Our approach focused on three complementary prompt strategies, progressively refined through experimentation:

\paragraph{P1 – Two-step pipeline strategy:} This strategy uses two consecutive prompts. The first phase performs a structural reduction by removing non-essential content, such as metadata, secondary names, and repetitive introductions, without altering the writing style. The second phase rewrites the remaining content using PL rules: short sentences, direct style, no complex subordinate clauses or repetitions, and simplified vocabulary. This approach was motivated by early observations of the training data, where many adaptations showed excessive content reduction. This two-step structure was designed to control that trend, especially in longer texts.

\paragraph{P2 – Unified prompt strategy:} This strategy applies a single prompt that combines reduction and rewriting instructions. The model processes the full text using rules for clarity, conciseness, and direct language. Specifically, it includes instructions to avoid technical terms, long sentences, and direct quotations. Examples of these instructions include:

\begin{itemize}
  \item Mant{\'e}n solo las personas e instituciones clave, como el alcalde, concejal de deportes u organizaci{\'o}n principal \textit{(Keep only key people and institutions, such as the mayor, sports councilor, or lead organization)}.
  \item Simplifica ubicaciones muy espec{\'i}ficas que no a{\~n}adan valor principal \textit{(Simplify very specific locations that do not add primary value)}.
  \item Resume listas de nombres o participantes secundarios utilizando expresiones como ``otras autoridades'' o ``otros participantes'' \textit{(Summarize lists of names or secondary participants using expressions such as ``other authorities'' or ``other participants'')}.
  \item Mant{\'e}n las ideas principales y conserva el vocabulario clave del original para asegurar alta coincidencia de palabras \textit{(Keep the main ideas and retain key vocabulary from the original to ensure high word matching)}.
\end{itemize}
This strategy aims to improve readability within acceptable ranges without fragmenting or omitting essential content.

\paragraph{P3 – Category-based strategy:} This strategy uses automatic classification to assign the input text to one of four categories: event listings, economic notices, short news items, or institutional notes. A set of common rewriting rules is applied, followed by category-specific adjustments such as formatting dates and prices, refining titles, or controlling the use of connectors. This classifier-based approach ensures stylistic and structural consistency depending on the content type.

\section{Experimental Setup and Evaluation}\label{sec:experimentalresults}

This section presents the experimental setup and the evaluation results of the systems described in Section~\ref{sec:architecture}. We report the performance of the two proposed system variants across both internal validation sets and the official evaluation data from Subtask~1. Evaluation focused on two metrics specified by the shared task organizers: Semantic Similarity and the Fern{\'a}ndez-Huerta Readability Index, as outlined in Section~\ref{sec:architecture}.

\subsection{Experimental Setup}\label{subsec:exp_settings}

All experiments were conducted on a shared Linux server running Ubuntu 22.04. The hardware setup included one NVIDIA L4 GPU (23\,GB VRAM), an Intel Xeon Silver processor, 64\,GB of RAM, and 877\,GB of NVMe SSD storage. The NVIDIA driver version was 535.230.02. All model runs were performed on GPU. The software environment is summarized in Table~\ref{tab:software_environment}.

\begin{table*}
  \caption{Software environment}
  \label{tab:software_environment}
  \begin{tabular}{ll}
    \toprule
    \textbf{Component} & \textbf{Version / Details} \\
    \midrule
    Python & 3.10.12 \\
    PyTorch & 2.6.0 + CUDA 12.4 \\
    Transformers (HuggingFace) & 4.50.0 \\
    PEFT & 0.15.1 \\
    CUDA Toolkit & 12.2 \\
    \bottomrule
  \end{tabular}
\end{table*}

\paragraph{System~1.} Salamandra-7B and RigoChat-7B-v2 in zero-shot mode, guided only by prompt engineering strategies P1, P2, and P3 were used. The generation process was configured with \texttt{do\_sample=False} to avoid sampling variability and promote output consistency. Input texts were first preprocessed using the normalization criteria described in Section~\ref{subsec:normalization}. After generation, outputs were postprocessed to correct residual errors and preserve structural clarity, as models occasionally reversed or ignored earlier preprocessing operations.

We ran prompt-based generation and fine-tuning experiments on three representative subsets of the training data ---\textit{smallest}, \textit{random}, and \textit{categories}--- to assess the effectiveness of each prompt strategy (P1, P2, P3) and identify the best-performing setup. These subsets, used for internal validation, are detailed in Section~\ref{subsec:subsets}.

\paragraph{System~2.} Fine-tuning of RigoChat-7B-v2 was performed using LoRA technique, allowing efficient training with minimal computational cost. We used the same training set composed of original texts and their PL adaptations, split into 85\% training and 15\% validation sets. Each fine-tuning experiment used a different prompt strategy (P1, P2, or P3), and all data were semantically normalized beforehand.

Fine-tuning was executed on an NVIDIA L4 GPU using the \texttt{transformers} and \texttt{PEFT} libraries. Training lasted approximately 2.5 hours per variant, with the following configuration: \texttt{r = 8}, \texttt{alpha = 32}, \texttt{dropout = 0.05}, \texttt{bias = "none"}, \texttt{task\_type = "CAUSAL\_LM"}, batch size = 16, evaluation at the end of each epoch, and between 3 and 5 epochs. All generated outputs were postprocessed to correct errors, ensure clarity, and remove non-essential content.

\subsection{Training Data Subsets}\label{subsec:subsets}

To evaluate the performance of the Salamandra and RigoChat models before the official test set was released, we created three reduced but representative subsets from the training data. This preliminary evaluation enabled us to refine our prompt strategies efficiently and select the best-performing configuration.

The subsets were designed with distinct selection criteria:

\begin{itemize}
  \item \textbf{Smallest subset:} Composed of 50 of the shortest texts in the dataset, characterized by low word count and brief structure (up to 90 words, with an average of 53.18 words). This subset was used to assess the structural simplification capabilities of the models.
  \item \textbf{Random subset:} Composed of 50 texts randomly sampled from the training set, representing a broad range of document lengths and topics (with an average word count of 447.78).
  \item \textbf{Categories subset:} Composed of 50 texts manually grouped by origin (e.g., Alicante, Elche, etc.), topic (e.g., Culture, Sport, etc.) and size (between 77 and 2,674 words, with an average of 818.34 words). This allowed us to explore how the models performed across different semantic domains.
\end{itemize}

These subsets were used for iterative prompt testing, qualitative review, and quantitative comparison, as shown in Section~\ref{subsec:internaleval}. This preliminary evaluation allowed us to calibrate our approach and ensure the reliability of the metrics applied to PL adaptations.

\subsection{Internal Evaluation on Training Subsets}\label{subsec:internaleval}

To evaluate the behavior of both models under different prompts and preprocessing conditions, we first conducted experiments on a subset of 150 texts (50 from each: \textit{smallest}, \textit{random}, and \textit{categories}). We tested System~1 in zero-shot mode using the Salamandra and RigoChat models combined with each prompt strategy (P1, P2, P3), applying the preprocessing and postprocessing pipeline described in Section~\ref{subsec:normalization}. 

We evaluated outputs using the official task metrics: Fern{\'a}ndez-Huerta for readability and Cosine Similarity for semantic preservation. The results are shown in Table~\ref{tab:internal-eval}.

\begin{table*}
  \centering
  \caption{Internal evaluation results (System~1) for Salamandra and RigoChat across prompt strategies}
  \label{tab:internal-eval}
  \begin{tabular}{lcccc}
    \toprule
    \textbf{Prompt} & \multicolumn{2}{c}{\textbf{Salamandra}} & \multicolumn{2}{c}{\textbf{RigoChat}} \\
    & FH & SIM & FH & SIM \\
    \midrule
    P1 & 74.57 & 0.69 & \textbf{76.80} & \textbf{0.81} \\
    P2 & 71.98 & 0.49 & \textbf{81.90} & \textbf{0.85} \\
    P3 & 83.22 & 0.50 & \textbf{84.75} & \textbf{0.80} \\
    \bottomrule
  \end{tabular}
\end{table*}

As shown in Table~\ref{tab:internal-eval}, RigoChat outperformed Salamandra in both metrics across all prompts, with the highest SIM from P2 (0.85) and best FH from P3 (84.75). Prompt P1 showed the weakest performance and was discarded.
Based on these results, we selected RigoChat for System~2 (fine-tuning) and retained only prompt strategies P2 and P3 for further experiments.

\begin{table*}
  \centering
  \caption{Performance scores on internal subsets (150 texts) using RigoChat (zero-shot) and RigoChat (fine-tuned)}
  \label{tab:finetune-eval}
  \begin{tabular}{lcccc}
    \toprule
    \textbf{Prompt} & \multicolumn{2}{c}{\textbf{RigoChat}} & \multicolumn{2}{c}{\textbf{RigoChat}} \\
     & \multicolumn{2}{c}{\textbf{(zero-shot)}} & \multicolumn{2}{c}{\textbf{(fine-tuned)}} \\
    & FH & SIM & FH & SIM \\
    \midrule
    P2 & 81.90 & \textbf{0.85} & \textbf{83.22} & \textbf{0.81} \\
    P3 & \textbf{84.75} & 0.80 & 83.33 & 0.78 \\
    \bottomrule
  \end{tabular}
\end{table*}

As shown in Table~\ref{tab:finetune-eval}, RigoChat with prompt P2 offered the best overall performance. It achieved the highest cosine similarity (0.85 in zero-shot, 0.81 after fine-tuning) and strong readability (FH = 83.22). Although P3 reached the best FH in zero-shot (84.75), it showed higher variability and reduced SIM after fine-tuning. This issue in P3 is likely due to heterogeneity in the training data ---differences in length, format, and structure that affected classification accuracy. P2, in contrast, consistently preserved key content while improving readability, without over simplifying or fragmenting the text. During the refinement phase, we noted that improving FH often required reformulation, which tended to reduce semantic similarity. P2 was the prompt to reach a stable equilibrium, preserving core content while raising readability to acceptable levels. 

Based on these findings, we selected RigoChat with prompt P2 and no fine-tuning (System~1) as our final configuration. Using this setup, we generated PL outputs for the 607 official test texts and submitted them to the CLEARS~2025 Subtask~1 evaluation.

\section{Results}\label{subsec:officialresults}

The official evaluation results for Subtask~1 are presented in Tables~\ref{tab:cosinesaverage} and~\ref{tab:fernandezhuerta}. Our system, HULAT-UC3M, ranked 1st in Semantic Similarity, achieving a score of 0.75, and 4th in readability, with a Fern{\'a}ndez-Huerta score of 69.72.

\begin{table*}
  \caption{Ranking based on Average Cosine Similarity}
  \label{tab:cosinesaverage}
  \begin{tabular}{llc}
    \toprule
    \textbf{Subtask~1} & \textbf{Team} & \textbf{Avg. Cosine Similarity} \\
    \midrule
    \textbf{1} & \textbf{HULAT-UC3M} & \textbf{0.75} \\
    2 & VICOMTECH & 0.71 \\
    3 & NIL\_UCM and CARDIFFNLP & 0.70 \\
    \bottomrule
  \end{tabular}
\end{table*}

\begin{table*}
  \caption{Ranking based on Average Fern{\'a}ndez-Huerta readability index}
  \label{tab:fernandezhuerta}
  \begin{tabular}{llc}
    \toprule
    \textbf{Subtask~1} & \textbf{Team} & \textbf{Avg. Fern{\'a}ndez-Huerta index} \\
    \midrule
    1 & VICOMTECH & \textbf{82.98} \\
    2 & CARDIFFNLP & 78.87 \\
    3 & NIL\_UCM & 70.42 \\
    \textbf{4} & \textbf{HULAT-UC3M} & 69.72 \\
    \bottomrule
  \end{tabular}
\end{table*}

Our system achieved the best result in semantic preservation, reflecting its capacity to retain core content meaning. Although it ranked last in FH readability, this outcome stemmed from a strategic decision. Preliminary analysis revealed that the training data contained limited reformulation and focused on deletion or surface-level simplification, diverging from the normative principles of ISO~24495-1:2023 (Plain Language).

Applying our standard reformulation methods would have produced clearer outputs but at the cost of lower similarity scores due to misalignment with the reference. To balance both dimensions, we chose an intermediate prompt strategy (P2), which maintains full content coverage while applying minimal, clarity-oriented changes. Since the evaluation protocol did not specify how SIM and FH would be weighted, we prioritized content fidelity and semantic completeness.

It is also worth noting that our FH score of 69.72 lies within the ``normal to somewhat easy'' readability range according to the FH scale, corresponding to secondary education (ESO 2-3) or B1 CEFR level-acceptable for general public communications with functional literacy.

\section{Error Analysis}\label{sec:erroranalysis}

This section presents an error analysis focused on two aspects: (1) limitations in the task-provided resources and (2) issues observed in our system's output.

\paragraph{Task-related limitations and insights.}
Several issues were identified during our participation in the task:

\begin{itemize}
  \item \textbf{Low semantic similarity in training pairs:} Initial inspection showed that many (original, PL) pairs in the training dataset had moderate-to-low semantic similarity, likely due to poor segmentation and unclear alignment. Some adaptations included additional content (e.g., entry ID=2346) or omitted elements from the original (e.g., ID=1075), which may have affected semantic coherence.
  \item \textbf{Impact on system performance:} These inconsistencies affected system behavior. In \textbf{System 2 (fine-tuning)}, conflicting examples limited the model’s ability to learn consistent simplification patterns. In \textbf{System 1 (zero-shot)}, the model sometimes under- or overgenerated content depending on the prompt, with limited reformulation, which reduced readability from very good to moderate-good levels.
  \item \textbf{Lack of adaptation methodology:} No adaptation methodology or PL guidelines were provided with the dataset. As a result, we had to adjust our standards-based approach. This was evident in some of the results; the lack of consistent simplification guidelines introduced variability that weakened model generalization.
\end{itemize}

\paragraph{Reflections on our own approach.} Within the constraints of the task, we identified areas in our system that could be improved:

\begin{itemize}
  \item Managing the trade-off between readability and similarity remained challenging. Enhancing clarity often required reformulation, which decreased cosine similarity, especially in the absence of reference adaptations aligned with ISO 24495-1:2023.    
  \item A more structured pipeline could help address this issue by incorporating text classification based on complexity, length, or target audience before generating simplifications based on LLM. This would allow models to adapt their behavior to specific input types and user needs, improving both clarity and content preservation.
\end{itemize}

\section{Conclusions}\label{sec:conclusions}

The organization of tasks like this one fosters research on Plain Language (PL) adaptation, an increasingly relevant topic for accessibility in domains such as healthcare and public services. This initiative contributes to the development of systems that support people with cognitive impairments and reading comprehension difficulties.

Despite the complexity of the task, our system achieved competitive results and yielded valuable insights into the challenges of PL adaptation. Our approach combined two complementary strategies: System~1, based on zero-shot inference with large language models (LLMs) using three prompt variants (P1, P2, P3); and System~2, a fine-tuned model trained on the task-provided data. The system finally selected was System~1. Prompt P2 was ultimately chosen for generation due to its robust and generalizable behavior across diverse input types. Unlike Prompt P3, which classified the input text by type (e.g., informative, event, and others) before applying specific strategies, often resulting in rigid outputs, P2 followed a simpler, generic instruction aimed at enhancing comprehension while preserving meaning. This balance, System~1 with prompt P2, proved effective, as deeper reformulations to optimize the Fern{\'a}ndez-Huerta index (readability) would have decreased similarity scores due to misalignment with reference texts. Given that the evaluation protocol did not specify the weighting between readability and similarity, we prioritized semantic preservation.

It would be advisable for future shared tasks to define a clear annotation methodology aligned with the regulatory context. Given the growing importance of PL in European and Spanish legislation, adopting standards like ISO 24495-1:2023 would help reduce variability and improve data quality.

Concerning evaluation, metrics remain a known limitation in the NLP community. Cosine similarity and the Fern{\'a}ndez-Huerta index capture only surface-level features and may penalize meaningful reformulation. Future frameworks should combine metric-based and user-centered approaches, covering aspects such as factual consistency, sentence restructuring, and lexical simplification, among others.

Looking ahead, we plan to obtain and analyze the adapted versions of the task test dataset to conduct a more targeted error analysis. Future work will explore approaches that not only consider the user, but actively involve them in the process. We also aim to design richer evaluation frameworks that combine automatic metrics with user validation, ensuring alignment with real-world accessibility needs and regulatory standards such as ISO 24495-1:2023 for Plain Language and UNE 153101:2018 EX for Easy-to-Read.

\begin{acknowledgments}
This work has been supported by grant PID2023-148577OB-C21 (Human-Centered AI: User-Driven Adapted Language Models-HUMAN\_AI) by MICIU/AEI/ 10.13039/501100011033 and by FEDER/UE.
\end{acknowledgments}

\bibliography{references}

\appendix

\section{Appendix - Examples of Inconsistencies in Training Dataset}

\begin{table*}
  \centering
  \caption{Examples of Inconsistencies in Training Data}
  \label{tab:dataset-examples}
  \begin{tabular}{|c|p{7cm}|p{7cm}|}
  \hline
  \textbf{ID} & \textbf{Original Text} & \textbf{PL Adaptation} \\
  \hline
2346 & 
``La Concejal{\'i}a de Igualdad les invita a participar en el programa de actividades ``Igualdad y diversidad en los barrios'' durante los meses de octubre a diciembre de 2022.

Si desea m{\'a}s informaci{\'o}n consulte los documentos en .pdf anexos al final de la noticia.'' 
& 
``La concejal{\'i}a de Igualdad presenta el programa de actividades de oto{\~n}o del 2022.

La concejal{\'i}a de Igualdad invita a todas las personas a participar en el programa de actividades de igualdad y diversidad en los barrios de Alicante de octubre a diciembre del 2022.

Programaci{\'o}n

Cuentacuentos  
Mili, Reportera superstar y los 3 cerditos:  
- El mi{\'e}rcoles 26 de octubre en la biblioteca Florida Babel  
- El mi{\'e}rcoles 23 de noviembre en la biblioteca El cabo  
- El lunes 12 de diciembre en la biblioteca Virgen del Remedio  

La verdad de las princesas:  
- El lunes 17 de octubre en la biblioteca Diagonal  
- El martes 18 de octubre en la biblioteca Villafranqueza  
- El mi{\'e}rcoles 19 de octubre en la biblioteca San Blas  

Malena quiere volar:  
- El mi{\'e}rcoles 5 de octubre en la biblioteca Benal{\'u}a  
- El lunes 10 de octubre en la biblioteca  
- El mi{\'e}rcoles 9 de noviembre en la biblioteca Diagonal  

¡Me lo pido!  
- El lunes 14 de noviembre en la biblioteca Virgen del Remedio  
- El martes 22 de noviembre en la biblioteca Carolinas  
- El mi{\'e}rcoles 30 de noviembre en la biblioteca Benal{\'u}a  

...

\textbf{[NOTE: Over 277 words words omitted due to space constraints]} \\
  \hline
1075 & 
``El alcalde valora la trayectoria personal y profesional del futbolista de Sanl{\'u}car de Barrameda, su capacidad goleadora y su intuici{\'o}n para situarse en el {\'a}rea del equipo contrario. Luis Barcala ha afirmado este domingo en el Sal{\'o}n Azul del Ayuntamiento de Alicante que ``este emblem{\'a}tico lugar ha sido testigo de recepciones a personalidades de distintos {\'a}mbitos de la vida pol{\'i}tica, social y cultural. Tambi{\'e}n los Alicantinos de Adopci{\'o}n, lo han visitado. Lo que no ha ocurrido nunca es que en {\'e}l se haya homenajeado a una leyenda, en blanco y azul, como es el caso de Eduardo Rodr{\'i}guez''.

Barcala ha agradecido a la Federaci{\'o}n de Casas Regionales en Alicante, que preside Miguel Beano, que haya pensado en Rodr{\'i}guez como Alicantino de Adopci{\'o}n 2022 ``en el a{\~n}o del centenario del H{\'e}rcules, y como reconocimiento al m{\'a}ximo goleador de su historia''.

El alcalde ha recordado c{\'c}mo lleg{\'o} Rodr{\'i}guez al H{\'e}rcules, procedente del Badajoz, pagando un traspaso de ocho millones. Ha subrayado, dirigi{\'e}ndose al homenajeado, que ``si eres hombre de pocas palabras, s{\'i} hablabas en el campo. Tu lenguaje era el de los goles''.

...

\textbf{[NOTE: Over 227 words omitted due to space constraints]} 
& 
``La Federaci{\'o}n de Casas Regionales en Alicante elige al futbolista Eduardo Rodr{\'i}guez como Alicantino de Adopci{\'o}n 2022. El alcalde de Alicante destac{\'o} en el acto de celebraci{\'o}n, la capacidad del futbolista del H{\'e}rcules para marcar goles y su intuici{\'o}n en el {\'a}rea del rival. Es la primera vez que se hace un reconocimiento a una leyenda blanca y azul y m{\'a}ximo goleador de la historia del club. El alcalde le felicit{\'o} por el reconocimiento como Alicantino en Adopci{\'o}n y por su carrera profesional. Al acto acudieron representantes de las Casas Regionales en Alicante, el ex capit{\'a}n del H{\'e}rcules Paquito y el ex presidente de la Diputaci{\'o}n Provincial. El acto se celebr{\'o} en el Sal{\'o}n Azul.'' \\
  \hline
  \end{tabular}
\end{table*}

\end{document}